\documentclass{article}
\usepackage{amsmath}
\usepackage{microtype}

\usepackage{makecell}
\usepackage{PRIMEarxiv}
\usepackage{graphicx}
\usepackage[utf8]{inputenc} 
\usepackage[T1]{fontenc}    
\usepackage{hyperref}       
\usepackage{url}            
\usepackage{booktabs}       
\usepackage{amsfonts}       
\usepackage{nicefrac}       
\usepackage{microtype}      
\usepackage{lipsum}
\usepackage{fancyhdr}       
\usepackage{graphicx}       
\graphicspath{{media/}}     

\title{Out-of-distribution detection based on subspace projection of high-dimensional features output by the last convolutional layer
}

\author{
  Qiuyu Zhu \\
  ShangHai University \\
  ShangHai\\
  \texttt{zhuqiuyu@staff.shu.edu.cn} \\
   \And
  Yiwei He \\
  ShangHai University \\
  ShangHai\\
  \texttt{heyiwei@shu.edu.cn} \\
}

\begin{document}
\maketitle

\begin{abstract}
Out-of-distribution (OOD) detection, crucial for reliable pattern classification, discerns whether a sample originates outside the training distribution. This paper concentrates on the high-dimensional features output by the final convolutional layer, which contain rich image features. Our key idea is to project these high-dimensional features into two specific feature subspaces, leveraging the dimensionality reduction capacity of the network's linear layers, trained with Predefined Evenly-Distribution Class Centroids (PEDCC)-Loss. This involves calculating the cosines of three projection angles and the norm values of features, thereby identifying distinctive information for in-distribution (ID) and OOD data, which assists in OOD detection. Building upon this, we have modified the batch normalization (BN) and ReLU layer preceding the fully connected layer, diminishing their impact on the output feature distributions and thereby widening the distribution gap between ID and OOD data features.  Our method requires only the training of the classification network model, eschewing any need for input pre-processing or specific OOD data pre-tuning. Extensive experiments on several benchmark datasets demonstrates that our approach delivers state-of-the-art performance. Our code is available at https://github.com/Hewell0/ProjOOD
\end{abstract}

\keywords{Out-of-distribution detection\and Convolutional layer\and Pattern recognition \and Subspace projection }

\section{Introduction}
Traditional deep learning models assume that training and testing datasets satisfy the independent and identically distributed condition. However, in real-world scenarios, inputs may originate from a data subspace distribution that significantly differs from that of the closed training set \cite{gawlikowski2021survey}. Samples deviating significantly from the distribution of in-distribution (ID) data are termed out-of-distribution (OOD) data.  \cite{yang2021generalized}.OOD detection is the cornerstone of various safety-critical applications, including autonomous driving \cite{ma2020artificial}, medical diagnosis \cite{litjens2017survey}, intelligent manufacturing \cite{ahmad2022deep}, and situations that require continuous learning.

Previous research predominantly utilized category-related information output by neural networks as the scoring function, such as the maximum posterior probability of a class derived from Softmax loss.\cite{hendrycks2016baseline}. Our main concern is utilizing the high-dimensional features output by the intermediate layers of neural networks, aiming to identify feature information that effectively detects OOD data.
\begin{figure*}[!t]
\centering
\includegraphics[scale=.4]{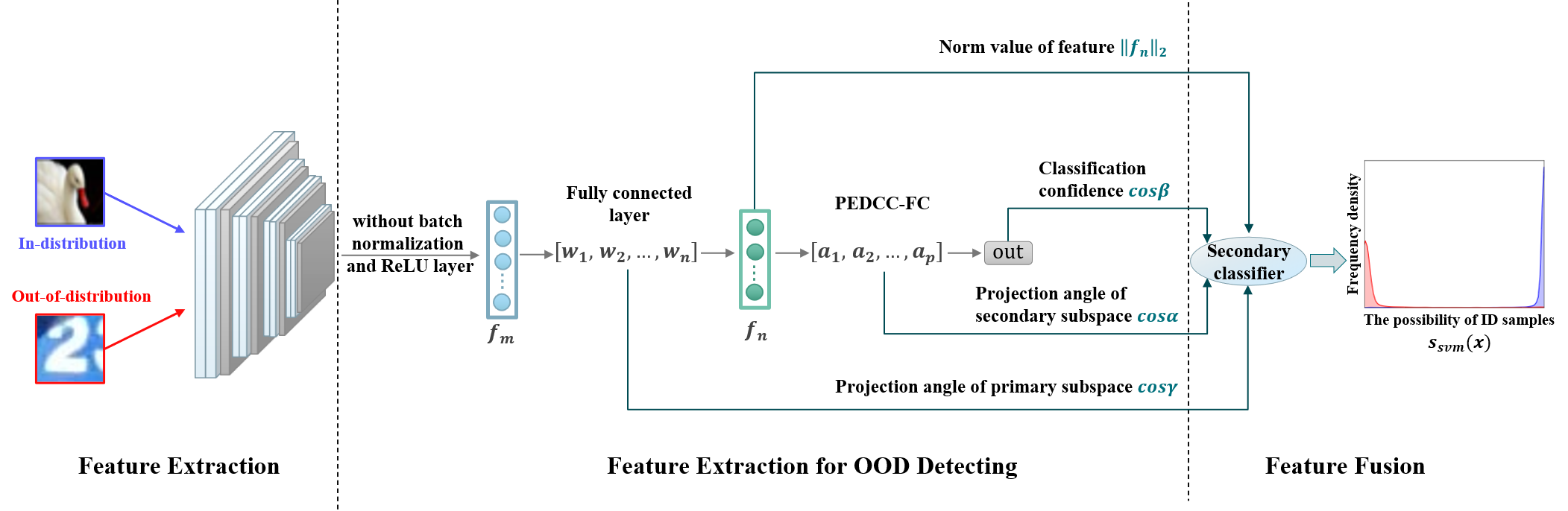}
\caption{Overview of the proposed pattern classification OOD detection framework based on the subspace projection of convolutional layer’s high-dimensional output features.}
\end{figure*}

During the training process, neural networks construct feature subspaces grounded in ID data. Previous studies indicates that, compared to general classifiers, the Predefined Evenly-Distribution Class Centroids (PEDCC)-Loss based classifier projects features into subspaces enriched with classification information\cite{zhu2019new}.Motivated by this, as illustrated in Figure 1, this paper extends the method and selects the outputs of the previous layer, which contain more ID data feature information. The primary subspace is projected by the fully connected layer after the last convolutional layer, which obtains a new projection angle. Feature-related metrics encompass a primary projection angle, along with two projection angles of the secondary PEDCC classification space and the norm values of the features.

Our method involves only canceling the BN and ReLU layers preceding the fully connected layer.  This approach does not add complexity to the model while preserving high classification accuracy on the ID dataset. Furthermore, our approach obviates the need for fine-tuning for each OOD dataset, aligning more closely with real-world scenarios. We conducted extensive evaluations of our method on benchmark image datasets, with experimental results demonstrating that the new scoring function, fusing multiple feature metrics, achieves state-of-the-art performance. The main contributions of this paper are summarized as follows:

\begin{enumerate}
\item A method for subspace projection, utilizing high-dimensional features extracted from the last convolutional layer of the neural network, is proposed.
\item An effective fusion method for OOD detection is employed to enhance the robustness of OOD detection.
\item Modifications to the BN and ReLU layers preceding the fully connected layer are made, thereby reducing constraints on the distribution of sample features.
\end{enumerate}

\section{Related work}
The essence of OOD detection lies in the binary classification of ID/OOD data \cite{yang2021generalized}.Upon inputting data $x$ into the model, the aim is to identify a scoring function $S(x)$ that quantifies classification uncertainty. The threshold $\tau$ of the binary classifier is then determined based on this scoring function.
\begin{equation}
isID(x)= \left\{ 
\begin{aligned}
True \quad S(x)\geq \tau, \\
False \quad S(x)< \tau.
\end{aligned} 
\right.
\end{equation}

\subsection{OOD detection based on softmax loss function and output information}
The selection of the scoring function S(x) determines the performance of OOD detection. \cite{hendrycks2016baseline} use the maximum posterior probability of the Softmax classifier as the scoring function, where ID data receive higher Softmax posterior probabilities than OOD data. The performance of using the Softmax posterior probability directly as the scoring function for OOD detection is limited. Based on this, \cite{liang2017enhancing} proposed ODIN using temperature scaling and input perturbation as two enhancement techniques for detection. By decomposing confidence scoring and a modified input pre-processing method, \cite{hsu2020generalized} further extended ODIN and designed a method called Generalized ODIN (G-ODIN), which achieved superior OOD detection results. \cite{huang2021mos} proposed a group-based OOD detection model that decomposes the large semantic space into smaller groups with similar concepts, and detected OOD data through the group Softmax loss.

In addition to utilizing the category Softmax probabilities, there is also other output information that can be used as a scoring function. \cite{liu2020energy} used the energy function of the logits for OOD detection. Base on energy function, \cite{zhang2022out} detect the OOD samples with Hopfield energy in a store-then-compare paradigm. \cite{sun2022dice}rank the weights by contribution and selectively use the most significant weights to enhance the detection ability of the energy function on OOD samples.

\subsection{OOD detection based on PEDCC-Loss}
As illustrated in Figure 2, our previously research uses the PEDCC classification cosine distance $\cos \theta _{i}$ and its decomposition angles: $\alpha$, which is the projection angle of feature $f_n$ onto the PEDCC framework; $\beta _{i}$, which is the angle between the projection $f_{p-sec}$ and the PEDCC framework \cite{hu2021generation}. We obtain their distributions, $S_\alpha (x)=\cos \alpha$, $S_\beta (x)=max\cos \beta _{i}$, achieving better OOD detection performance than G-ODIN.

\begin{figure}[!h]
\centering
\includegraphics[scale=.5]{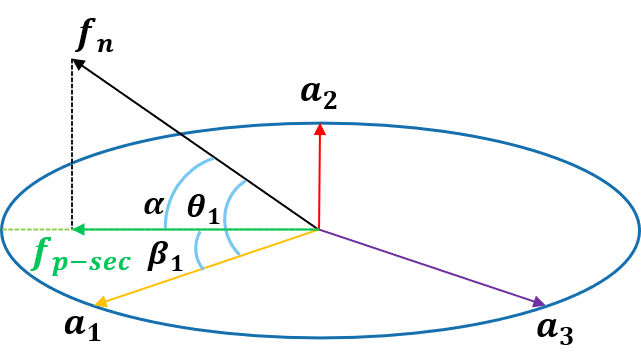}
\caption{Illustration of feature projection based on PEDCC in a three-dimensional space. The two-dimensional space is spanned by predefined evenly distributed unit vectors $a_1$, $a_2$, and $a_3$. }
\end{figure}

PEDCC-Loss \cite{zhu2019classification} is the loss function based on
PEDCC. PEDCC-Loss function is as follows:
\begin{equation}
L_{PEDCC-AM}=- \frac{1}{N}\sum _{i=1}^{N}\log \frac{e^{s \cdot(\cos \theta _{i}-m)}}{e^{s \cdot(\cos \theta _{i}-m)}+ \sum _{j=1,j\neq i}^{c}e^{s \cdot \cos \theta _{j}}}
\end{equation}
\begin{equation}
L_{PEDCC-MSE}= \frac{1}{N}\sum _{i=1}^{N}||f_{i}-a_{i}||^{2}
\end{equation}
\begin{equation}
L_{PEDCC}=L_{PEDCC-AM}+L_{PEDCC-MSE}
\end{equation}
$L_{PEDCC-AM}$ is the AM-Softmax loss\cite{Wang_2018}, $s$ and $m$ are adjustable parameters.$L_{PEDCC-MSE}$ is the MSE loss. The addition of the two is PEDCC-Loss, where $n (n \geq 1)$ is a constraint factor of the $L_{PEDCC-MSE}$.
\section{Method}
A well-trained model has a strong ability to extract semantic features from inputs. Upon the input of OOD data, the model can, to some extent, discern the differences between these inputs and ID data. This is reflected in the feature information output by the model's intermediate layer.
\subsection{The primary subspace projection}
The classification model based on PEDCC-Loss adapts the classification features to the feature space surrounding each class centroid through network training. Given that the dimensional requirement of the classification feature for PEDCC-Loss is lesser than that of the feature output by the last convolutional layer, a fully connected layer is employed to reduce the high-dimensional feature to a lower dimension.

\begin{figure}[!t]
\centering
\includegraphics[scale=.4]{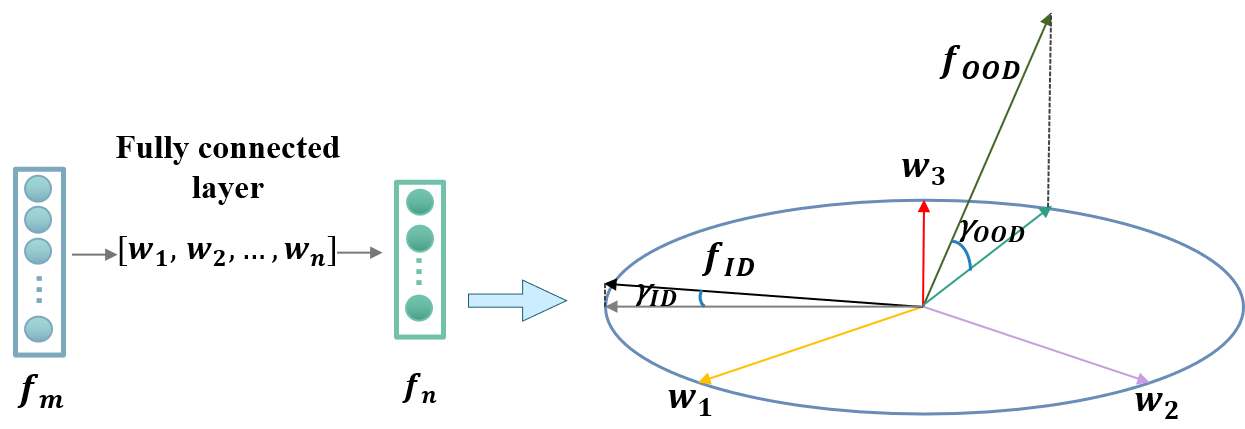}
\caption{Illustration of primary subspace projection by the first fully connected layer. The feature $f_m$ output by the last convolutional layer is projected onto the subspace spanned by the vectors $w_1, w_2,..., w_n$. }
\end{figure}

As illustrated in Figure 3, The input features $f_m$ can be considered as being projected onto a specific feature subspace, spanned by the weight matrix $W=[w_1, w_2,\cdots, w_n]$ of the first fully connected layer.The output dimension of the last convolutional layer is denoted as $m$. This dimension $m$ is larger than the output dimension $n$ of the first fully connected layer. The projection matrix P of the first fully connected layer’s weights can be obtained using the following formula:
\begin{equation}
P=W(W^{T}W)^{-1}W^{T}
\end{equation}
and then we get the projection $f_{p-pri}$ of $f_m$:
\begin{equation}
f_{p-pri}=Pf_{m}
\end{equation}

After a certain number of training iterations, the network projects the features output by the convolutional layers into a specific feature subspace, aiming to minimize the projection angle $\gamma$ between $f_m$ and $f_{p-pri}$. The projection subspace contains ID classification information. Since the network is not exposed to OOD data during training, it tends not to project them into the specific subspace. The projection angle between the OOD data's feature and the weight projection subspace tends to be larger than that of the ID data. Therefore, OOD data can be more effectively detected using the scoring function $S_{\gamma}(x)$.
\begin{equation}
S_{\gamma}(x)= \cos \gamma=\frac{f_{p-pri}}{f_{m}}
\end{equation}

The projection matrix must satisfy the condition of linear independence. If the weight matrix $W$ fails to achieve linear independence in $n$ dimensions, it may collapse into a lower-dimensional subspace, potentially increasing the error rate in detecting ID samples as OOD samples. Therefore, during network training, imposing constraints on the projection matrix to ensure its linear independence is essential. A linear-independence loss function is applied to eliminate linear dependencies and constrain the weights of the first fully connected layer.
\begin{equation}
L_{Lin-ind}= \frac{1}{n(n-1)}\sum _{i}^{n}\sum _{j \neq i}^{n}(W_{norm}\times W_{norm}^{T}-I_{n})
\end{equation}
\begin{equation}
W_{norm}= \frac{W}{||W||_{2}}
\end{equation}
where $I_n$ represents an n-dimensional diagonal identity matrix.

We use the combination of PEDCC-Loss and linear-independence loss as the loss function in classification task. $k$ are adjustable parameters.

\begin{equation}
L=L_{PEDCC}+k*L_{Lin-ind}
\end{equation}

For both the classification of ID data and the detection of OOD data, the model trained with the biased fully connected layer outperforms the unbiased model. However, considering the bias as a dimension in the projection subspace vectors adversely affects OOD detection effectiveness. This is because, in the classification training process, the fully connected layer's bias is tantamount to adding constants, lacking effective category information and ID/OOD data detection capability. Consequently, the weight matrix alone is used to span the projection subspace, while retaining the biased fully connected layer for training in the classification process.

\subsection{The norm value of feature}
The model trained with ID data has a strong feature extraction ability , enabling the salient features of ID data to activate the neural network maximally \cite{taigman2015web}. OOD data, lacking ID data's unique semantic features, result in a lower degree of neural network activation. The magnitude of the PEDCC feature output by the first fully connected layer represents the extent to which the model is activated by the feature. An increase in this magnitude signifies a higher degree of activation. To calculate the magnitude of the output feature, it suffices to compute the norm of the feature, which is straightforward and highly operable. The $L_{ 2 }$-norm is used to determine the norm value of the PEDCC feature output by the first fully connected layer, with the scoring function based on this feature norm value as follows:
\begin{equation}
S_{norm}(x)=||f_{n}||_{2}
\end{equation}
where $f_n$ represents the feature output by the first full connected layer. 

\subsection{Multi-feature fusion method}
Although a single feature metric can achieve a certain level of OOD detection performance, it may not be optimal. As the number of ID data categories increases, the network's ability to extract features for ID data diminishes, leading to reduced OOD detection performance. The single feature metric proves unsuitable for large-scale datasets with an excessive number of categories. The multi-feature fusion method effectively complements the useful information between multiple feature metrics. This method not only preserves OOD detection performance but also enhances the algorithm's stability.

The fusion method adopted involves training the secondary classifier. Four feature metrics,$\cos \alpha$ , $\cos \beta _{i}$ 
, $\cos \gamma$ and $||f_{n}||_{2}$, are selected as the input features of the support vector machine (SVM). Probability estimates are utilized, employing the SVM's output probabilities as the scoring function:
\begin{equation}
S_{svm}(x)=prob_{out}
\end{equation}

\subsection{Constraints of The BN and ReLU Layer}
To enhance the classification accuracy of neural networks, Batch Normalization (BN) \cite{ioffe2015batch} and Rectified Linear Unit (ReLU) layers \cite{glorot2011deep} are added to the input of each layer. In classifying ID data, these operations expedite convergence but also reduce the feature distribution distances \cite{hein2019relu} , resulting in features of the same category clustering in the same feature space.

As shown in Figure 4,various operations preceding the fully connected layer differentially impact feature distributions. The unprocessed convolution layer's feature projection angle cosine value distribution exhibits a bimodal pattern with minimal overlap. 
After BN and ReLU layers application, the ID data's projection angle cosine value distribution becomes more concentrated, illustrating BN and ReLU's constraining effect on feature distribution in classification tasks. However, this simultaneously reduces the distance between OOD and ID feature distributions and increases the overlap at the center. This hinders the identification of a threshold to distinguish between them.
\begin{figure}[!h]
\centering
\includegraphics[scale=.5]{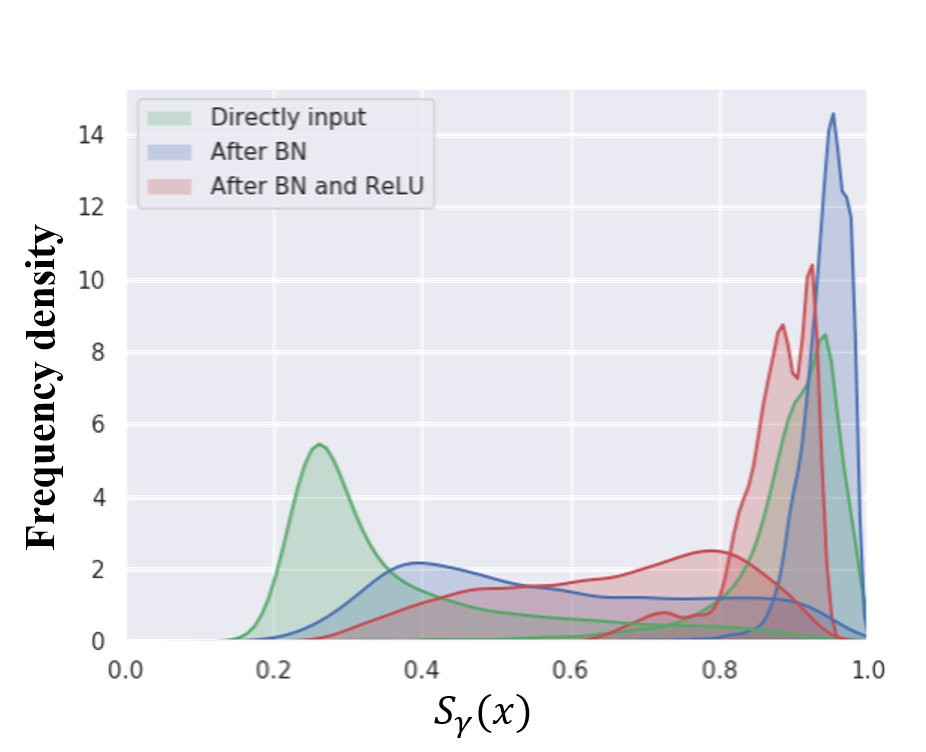}
\caption{Distributions of cosines of primary projection angles across different network structures. In these distributions, segments to the left of the same color represent OOD data, while those to the right indicate ID data.}
\end{figure}
\section{Experiments and discussion}
\subsection{Experimental settings}
\paragraph{Datasets}
We use SVHN and CIFAR-10/100 images with size 32X32 for the classification tasks. We include all the OOD dataset used in G-ODIN\cite{hsu2020generalized}, which are TinyImageNet(crop), TinyImageNet(resize)\cite{deng2009imagenet}, LSUN(crop), LSUN(resize)\cite{yu2015lsun}, iSUN, and SVHN.
\paragraph{Evaluation metrics}
We use the two most widely adopted metrics in the OOD detection and one newly proposed metric. The first one is the area under the receiver operating characteristic curve (AUROC). The second one is the true negative rate at 95$\%$ true positive rate (TNR at TPR 95$\%$) \cite{fawcett2006introduction}. In addition, to attain high accuracy in recognizing both ID and OOD samples simultaneously, we propose a new metric true negative rate at 98$\%$ true positive rate (TNR at TPR 98$\%$). 
\paragraph{Networks and training details}
DenseNet \cite{he2016deep} was used for the classifier backbone. DenseNet has 100 layers with a growth rate of 12. It was trained with a batch size of 128 for 100 epochs with weight decay of 0.0005. In both training, the optimizer is SGD with momentum 0.9, and the learning rate starts with 0.1 and decreases by factor 0.1 at 30$\%$ and 60$\%$ of the training epochs. The model parameters saved at the 70th epoch was used for OOD sample detection. In all experiments, TinyImageNet (resize) served as the reference OOD dataset, and classification training samples were extracted as the ID dataset to train the SVM, with a penalty parameter C set at 5. The parameters in $L_{PEDCC-AM}$ mentioned in 2.2 were selected based on the highest classification accuracy\cite{zhu2022effective}, with m=5.5 and s=0.35 for CIFAR-10 and SVHN,  m=10 and s=0.25 for CIFAR-100 in Eq.(2).k=1 for CIFAR-10 and SVHN,  k=100 for CIFAR-100 in Eq.(10).

\begin{table*}[!t]
\caption{\label{tab1}OOD detection performance comparison between ours and other five methods(All values are percentages. \textbf{Bold} numbers are superior results.)}
\centering
\resizebox{\linewidth}{!}{
\begin{tabular}{llccc}
    \toprule
    ID & OOD & TNR at TPR 95$\%$$\uparrow$ &AUROC$\uparrow$& TNR at TPR 98$\%$$\uparrow$\\
    \cmidrule(lr){3-5}
    {\quad} & {\quad}& \multicolumn{2}{c}  {Baseline/ODIN/Mahalanobis/G-ODIN/PEDCC/Ours}&Ours \\
    \midrule
    CIFAR-10  & TinyImagenet(c) & 50.0/47.8/81.2/93.4/98.3/\textbf{99.79} & 92.1/88.2/96.3/98.7/99.1/\textbf{99.58} &99.36\\
                    {\quad}& TinyImagenet(r) & 47.4/51.9/90.9/95.8/96.4/\textbf{99.82} & 91.5/90.1/98.2/99.1/99.2/\textbf{99.71}&99.57\\
                    {\quad}& LSUN(c) & 51.8/63.5/64.2/91.5/99.0/\textbf{99.99} &  93.0/91.3/92.2/98.3/99.6/\textbf{99.74}&99.94\\
                    {\quad}& LSUN(r) & 56.3/59.2/91.7/97.6/98.6/\textbf{99.95} &  93.9/92.9/98.2/99.4/99.2/\textbf{99.86}&99.87\\
                    {\quad}& iSUN	  & 52.3/57.2/90.6/97.5/98.9/\textbf{99.90} & 93.0/92.2/98.2/99.4/99.5/\textbf{99.87}&99,79\\
                    {\quad}& SVHN	  & 40.5/48.7/90.6/94.0/95.9/\textbf{99.88}	& 88.1/89.6/98.0/98.8/99.0/\textbf{99.65}&99.75\\
    CIFAR-100  & TinyImagenet(c) & 25.3/56.0/63.5/87.8/92.0/\textbf{98.24} & 79.0/90.5/92.4/97.6/98.5/\textbf{99.45}&95.60\\
                    {\quad}& TinyImagenet(r) & 22.3/59.4/82.0/93.3/88.4/\textbf{97.13} & 76.4/91.1/96.4/98.6/98.1/\textbf{99.21}&94.58\\
                    {\quad}& LSUN(c) & 23.0/53.0/31.6/75.0/\textbf{93.5}/91.68 &  78.6/89.9/81.2/95.3/\textbf{98.7}/97.66&83.34\\
                    {\quad}& LSUN(r) & 23.7/64.0/82.6/93.8/88.3/\textbf{99.36} &  78.2/93.0/96.6/98.7/97.4/\textbf{99.76}&98.46\\
                    {\quad}& iSUN	  & 21.5/58.4/81.2/92.5/88.9/\textbf{98.65} & 76.8/91.6/96.5/98.4/98.2/\textbf{99.58}&97.23\\
                    {\quad}& SVHN	  & 18.9/35.3/43.3/77.0/87.1/\textbf{99.54}	& 78.1/85.6/89.9/95.9/97.8/\textbf{99.73}&98.52\\
    SVHN  & TinyImagenet(c) & 81.3/88.5/88.6/-/97.6/\textbf{98.48} & 93.3/96.0/96.2/-/98.6/\textbf{99.60}&95.37\\
                    {\quad}& TinyImagenet(r) & 79.8/84.1/93.0/-/97.1/\textbf{99.77}& 94.8/95.1/98.1/-/98.5/\textbf{99.89}&99.13\\
                    {\quad}& LSUN(c) & 59.7/83.3/73.1/-/\textbf{93.2}/90.27&  89.5/93.8/94.6/-/97.1/\textbf{98.35}&84.69\\
                    {\quad}& LSUN(r) & 77.1/81.1/91.2/-/94.4/\textbf{99.99} &  94.1/94.5/93.7/-/98.2/\textbf{99.95}&99.64\\
                    {\quad}& iSUN	  & 73.8/93.5/92.8/-/96.3/\textbf{99.85} & 89.2/97.3/96.6/-/98.5/\textbf{99.91}&99.07\\
                    {\quad}& CIFAR-10	  & 69.3/71.7/57.7/-/92.3/\textbf{99.96} & 91.9/91.4/89.5/-/97.4/\textbf{99.95}&99.75\\
    \bottomrule
\end{tabular}
}
\end{table*}

\subsection{Results and discussion}
\subsubsection{OOD detecting performance}
Table 1 shows that an overall comparison for methods that train without using a specific OOD dataset for parameter tuning in advance with 6 OOD benchmark datasets. We choose 
Baseline\cite{hendrycks2016baseline}, ODIN\cite{liang2017enhancing}, Mahalanobis\cite{lee2018simple}, G-ODIN\cite{hsu2020generalized} and our previous work $S_{PEDCC}$\cite{zhu2022out} as the method of comparison. It can be noticed that our proposed approach can perform well in various datasets, and the TNR at TPR 95$\%$ 
is mostly close to 100$\%$, significantly better perform than other OOD detection methods. When the number of ID data categories increases, our method can also preserve high OOD detection performance, which shows that our proposed approach is very effective.In addition, on our proposed new metric TNR at TPR98$\%$, our method can also achieve a better detection level, which shows that we can simultaneously identify ID data and OOD data with high accuracy. This proves the features output by the convolution layer contain more feature information beneficial to OOD detection than the resulting confidence.

Our OOD detection method directly uses the classification-trained model, so it will not affect the classification accuracy of ID data. The classification accuracy of CIFAR10 is 92.7$\%$, CIFAR100 is 74.04$\%$, and SVHN is 95.93$\%$. Our method directly adopts the parameters of the classification network trained by PEDCC-Loss without additional training process or complex input preprocessing, only requiring an appropriate threshold like baseline. In other words, our algorithm has the same simplicity as the baseline, and achieves state-of-the-art performance.

\subsubsection{Ablation study}
\paragraph{The influence of feature projection angle and Feature norm on OOD detection}

We study the effect of the feature projection angle and feature norm mentioned in sections 3.1 and 3.2 on OOD detection, as shown in Table 2, $S_{\gamma}(x)$ and $S_{norm} (x)$ have a certain detection ability for OOD data. It shows that the OOD detection effectiveness of network intermediate features. However, with the increase of ID data categories, the detection performance of a single feature metric will also decrease to a certain extent. It reflects that the advantage of the multi-feature fusion method is to improve the detection performance in situations where there are a large number of categories in ID data. The multi-feature fusion method greatly assists in enhancing the detection's robustness.
\begin{table*}[!t]
\caption{\label{tab2}Detection performance of Different training strategies }
\centering
\resizebox{\linewidth}{!}{
\begin{tabular}{llcc}
    \toprule
    ID& OOD & \makecell[c]{TNR at \\ TPR 95$\%$$\uparrow$ }&{AUROC$\uparrow$}\\
    \cmidrule(lr){3-4}
    {\quad} & {\quad}& \multicolumn{2}{c}  {$S_{\gamma}(x)$ / $S_{norm} (x)$/$SVM_{\alpha\&\beta}$ / $SVM_{\gamma\&norm}$/Logistic regression/Linear kernel/ Gaussian kernel} \\
    \midrule
    CIFAR-10  & TinyImagenet(r) & 81.92/73.83/69.06/99.86/83.82/83.86/99.82 & 96.59/94.88/94.87/99.95/97.28/97.06/99.71\\
                    {\quad}& LSUN(r) & 88.83/82.17/72.33/99.72/91.05/90.73/99.95 & 97.93/96.49/95.68/99.99/98.46/98.16/99.86\\
                    {\quad}& iSUN	 & 89.49/83.44/72.92/99.65/92.01/91.49/99.90 & 98.11/96.65/95.52/99.98/98.53/98.19/99.87\\
    CIFAR-100  &  TinyImagenet(r) & 65.94/12.45/44.25/84.84/80.32/81.02/97.13 & 93.31/76.90/88.54/96.39/96.26/96.36/99.21\\
                    {\quad}& LSUN(r) &64.29/10.85/56.31/92.40/89.21/90.29/99.36 & 93.50/78.90/91.29/98.37/97.81/97.93/99.76\\
                    {\quad}& iSUN	  & 65.37/10.75/53.96/86.35/87.09/87.96/98.65 & 93.48/77.95/90.62/96.76/97.61/97.71/99.58\\
    SVHN  & TinyImagenet(r) & 86.44/80.17/82.64/99.67/86.65/87.36/99.77 & 96.96/95.99/99.72/99.88/97.87/97.87/99.89\\
                    {\quad}& LSUN(r) &85.02/76.71/80.74/99.99/86.54/86.76/99.99 & 96.67/95.37/96.41/99.95/97.97/97.93/99.95\\
                    {\quad}& iSUN	  & 87.50/80.39/83.54/99.76/87.93/88.42/99.85 & 97.15/96.05/96.70/99.91/98.10/98.08/99.91\\
    \bottomrule
\end{tabular}
}
\end{table*}

\paragraph{Selection of feature metrics}

As shown in Table 2, we experimented different selections of the input feature metrics of the SVM,$SVM_{\alpha\&\beta}$ represents SVM that uses $\cos \alpha$ and $max\cos \beta _{i}$ as input features, and $SVM_{\gamma\&norm}$  represents SVM that uses $\cos \gamma$ and $||f_{n}||_{2}$ as input features. We found $SVM_{\gamma\&norm}$ has better OOD detection performance than $SVM_{\alpha\&\beta}$. The $\cos \alpha$ and $max\cos \beta _{i}$ are obtained by decomposing the cosine distance of the PEDCC output, which also relates to the result information output by the neural network. $SVM_{\gamma\&norm}$  has excellent performance in some datasets. We argue that the output features of the intermediate layer of the neural network contain ID specific feature information, which is more instructive for OOD sample detection than the final output of the network.

\begin{table}[htbp]
\caption{\label{tab5} Comparison of detection performance of different reference OOD datasets}
\centering
\begin{tabular}{cccc}
    \toprule
     OOD & \multicolumn{3}{c}  {TNR at TPR 95$\%$$\uparrow$ }\\
    \cmidrule(lr){2-4}
    {\quad} & \makecell[c]{Tuning with\\TinyImagenet(r)}& \makecell[c]{Tuning with\\LSUN(r)} &{Tuning Each}  \\
    \midrule
    \makecell[c]{Tiny\\Imagenet(c)} & 99.79&99.51&99.83\\
    \makecell[c]{Tiny\\Imagenet(r)}	& 99.82&99.72&99.82\\
    LSUN(c) & 99.99&99.99&99.99\\
    LSUN(r) & 99.95&99.96&99.96\\
    iSUN & 99.90&99.96&99.96\\
    SVHN & 99.88&99.85&99.89\\

    \bottomrule
\end{tabular}
\end{table}
\subsubsection{Selection of multi-feature fusion methods}
We consider that the sample composed of four metrics of $\cos \alpha$ , $max\cos \beta _{i}$, $\cos \gamma$ and $||f_{n}||_{2}$ are not necessarily linearly separable, so we try to use a nonlinear secondary classifier model. In the experiment, after using min-max standardization on the feature values obtained in the network, we choose the SVM with Gaussian kernel as the secondary classifier.

Table 2 shows the experimental results of feature fusion using linear logistic regression and using linear kernel SVM. The linear logistic regression model employed a regularization parameter of $L_2$, a penalty value of 0.5 for the loss function, and a gradient descent method for solution. The settings of linear kernel SVM are consistent with Gaussian kernel SVM. Finally, it is found that the SVM using the Gaussian kernel can achieve the best detection performance.

To circumvent the need for fine-tuning the secondary classifier for each OOD dataset, the strategy involves randomly selecting an OOD dataset as training data for the SVM, and subsequently applying the trained SVM to various other OOD datasets. This necessitates that the secondary classifier possesses robust generalization capabilities. The experimental findings demonstrate that the distribution of feature metrics across diverse OOD datasets is approximately equivalent. Consequently, it is feasible to employ a non-specific OOD dataset to train the secondary classifier.

Table 3 shows that the detection performances of the secondary classifier tuning with TinyImagenet(r),  tuning with another reference OOD dataset LSUN(r) and tuning with each dataset. Other settings are the same as the previous experiments. The detection performance of different SVM tuning methods is basically equal, and can reach the optimal level. The results of tuning with each dataset have a small improvement range compared with the results of tuning with one dataset. The experimental results also show that the selection of OOD dataset will not affect the performance of the detection method. Our method does not need to use specific OOD dataset to tune secondary classifier in advance, and make it more analogous to the reality of the open world.

\begin{table}[htbp]
\caption{\label{tab9} Comparison of detection performance of $S_{\gamma}(x)$ with different bias Settings when the ID dataset is CIFAR-10 (Using means using bias as the one-dimensional vector of the projection matrix)}
\centering
 \resizebox{0.5\textwidth}{!}{
\begin{tabular}{cccc}
    \toprule
     OOD & \multicolumn{3}{c}  {TNR at TPR 95$\%$$\uparrow$ }\\
    \cmidrule(lr){2-4}
    {\quad} & \makecell[c]{Training \\without bias} & \makecell[c]{Training with bias\\and Using} & \makecell[c]{Training with bias\\but not Using}\\
    \midrule
    \makecell[c]{Tiny\\Imagenet(c)} & 76.46 &77.89&84.24\\
    \makecell[c]{Tiny\\Imagenet(r)}	& 83.33 &78.86&81.92\\
    LSUN(c) & 89.37&89.91&92.08\\
    LSUN(r) & 85.11&85.80&88.83\\
    iSUN & 88.47&86.72&89.49\\
    SVHN & 93.42&94.24&95.43\\

    \bottomrule
\end{tabular}}
\end{table}

\subsubsection{Effect of bias in fully connected layer}
We experimentally compare the effect of bias. The classification results are 92.7$\%$ for the model trained with bias and 92.4$\%$ for the model trained without bias. The influence of bias on OOD detection results is shown in Table 4. Since the model trained with bias performs better in ID data classification, the ID data feature projection angle will also be smaller, and the detection discrimination of the scoring function $S_{\gamma}(x)$ is higher. In training, the bias is equivalent to a constant, which does not contain the category knowledge of the ID data. If the bias is used as a one-dimensional vector of the projection matrix, the category feature information of ID data will be blurred, and the detection accuracy of OOD data will be reduced. Therefore, in our method, the weight matrix of the fully connected layer with bias training is selected as the subspace projection matrix.
\begin{table}[htbp]
\caption{\label{tab8}Comparison of detection performance of different network structures}
\centering
\begin{tabular}{ccc}
    \toprule
     OOD & \multicolumn{2}{c}  {TNR at TPR 95$\%$$\uparrow$ }\\
    \cmidrule(lr){2-3}
    {\quad} & {Post BN and ReLU before} & {Directly input }\\
    \midrule
    \makecell[c]{Tiny\\Imagenet(c)} & 71.75&99.79 \\
    \makecell[c]{Tiny\\Imagenet(r)}	& 93.61&99.82\\
    LSUN(c) & 35.61&99.99\\
    LSUN(r) & 95.74&99.95\\
    iSUN & 93.48&99.90\\
    SVHN & 94.31&99.88\\

    \bottomrule
\end{tabular}
\end{table}
\subsubsection{Effect of the BN and ReLU layer before fully connected layer}
We experimentally compare two network structures, one uses the BatchNorm2d function and ReLU function provided in the torch.nn module to activate the high-dimensional features output by the last convolutional layer, and the other directly inputs high-dimensional features into the fully connected layer. Other settings are the same as the previous experiments. The experimental results are shown in Table 5.The network with the final BN and ReLU layer cannot detect OOD data well. It can be concluded that the high-dimensional features are directly input to the fully connected layer, which can reduce the constraint effect of the BN and ReLU layer on the feature distribution. OOD data features will not be approximated to the feature distribution of ID data, and the distance between the two feature distributions is maintained, which is helpful for OOD detection.

\section{Conclusion}
In this paper, for the OOD detection of deep learning pattern recognition, we propose an OOD detection method based on network intermediate features output by the last convolutional layer. We increase the feature distribution distance between OOD data and ID data by modifying the BN and ReLU layers. Our method is based on the dimensionality reduction property of the network's linear layers with the PEDCC-Loss. We employ two levels of subspace projection of the high-dimensional features output by the convolution layer. The cosines of the three projection angles, along with the norm value of the feature, are computed, encapsulating the unique feature information of the In-Distribution data. Subsequently, a secondary classifier for OOD detection is devised, fusing multiple feature metrics without requiring additional tuning for each specific dataset. Extensive experiments demonstrate that our proposed approach has achieved state-of-the-art performance. Through multi-feature fusion, our method exhibits high robustness on a large-scale dataset. In future work, we plan to use a large-scale pre-trained model to apply the OOD detection method based on intermediate features to even larger-scale datasets.
\bibliographystyle{unsrt}  
\bibliography{references}

\end{document}